\documentclass[journal]{IEEEtran}

\usepackage{amsmath,epsfig}
\usepackage{subfigure}
\usepackage{array}
\usepackage{framed,multirow}

\usepackage[colorlinks,linkcolor=red,hyperindex,bookmarks=false]{hyperref}
\usepackage{cite}

\begin{document}

\title{An $\alpha$-Matte Boundary Defocus Model Based Cascaded Network for Multi-focus Image Fusion}

\author{Haoyu Ma, Qingmin Liao, Juncheng Zhang, Shaojun Liu, Jing-Hao Xue%
\thanks{This work was partly supported by the National Natural Science Foundation of China (61771276) and the National Key Research and Development Program of China (2016YFB0101001). This work uses a part of the Adobe Deep Matting dataset \cite{xu2017deep} for model training. (Corresponding author: Jing-Hao Xue.)}%
\thanks{H.~Ma, Q.~Liao and J.~Zhang are with the Department of Electronic Engineering, Tsinghua University, China. Email: hy-ma17@mails.tsinghua.edu.cn, liaoqm@tsinghua.edu.cn, zjc16@mails.tsinghua.edu.cn}
\thanks{S.~Liu is with the Department of Electronic and Computer Engineering, Hong Kong University of Science and Technology, China. Email: liusj14@tsinghua.org.cn}
\thanks{J.-H.~Xue is with the Department of Statistical Science, University College London, UK. Email: jinghao.xue@ucl.ac.uk}
\thanks{Manuscript received October 28, 2019; revised .}}

\markboth{IEEE Transactions on Image Processing}%
{Ma \MakeLowercase{\textit{et al.}}: MMF-Net}

\maketitle

\begin{abstract}
Capturing an all-in-focus image with a single camera is difficult since the depth of field of the camera is usually limited. An alternative method to obtain the all-in-focus image is to fuse several images focusing at different depths. However, existing multi-focus image fusion methods cannot obtain clear results for areas near the focused/defocused boundary (FDB). In this paper, a novel $\alpha$-matte boundary defocus model is proposed to generate realistic training data with the defocus spread effect precisely modeled, especially for areas near the FDB. Based on this $\alpha$-matte defocus model and the generated data, a cascaded boundary aware convolutional network termed MMF-Net is proposed and trained, aiming to achieve clearer fusion results around the FDB. More specifically, the MMF-Net consists of two cascaded sub-nets for initial fusion and boundary fusion, respectively; these two sub-nets are designed to first obtain a guidance map of FDB and then refine the fusion near the FDB. Experiments demonstrate that with the help of the new $\alpha$-matte boundary defocus model, the proposed MMF-Net outperforms the state-of-the-art methods both qualitatively and quantitatively.
\end{abstract}


\IEEEpeerreviewmaketitle

\section{Introduction}\label{s:intro}

\begin{figure}[htbp]
\centering 
  \subfigure[Source image A]{\label{fig:subfig:11}
    \includegraphics[width=0.32\linewidth]{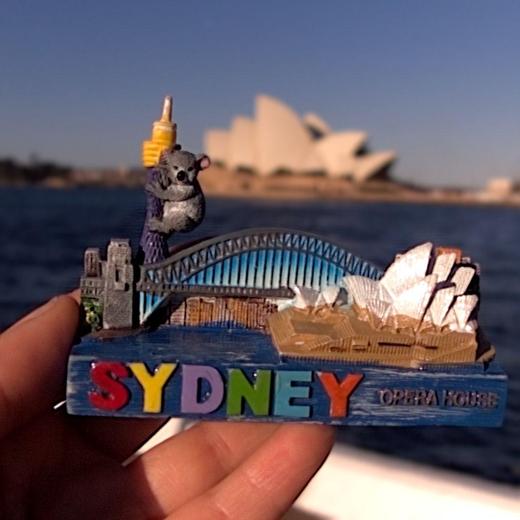}}%
  \subfigure[Source image B]{\label{fig:subfig:12}
    \includegraphics[width=0.32\linewidth]{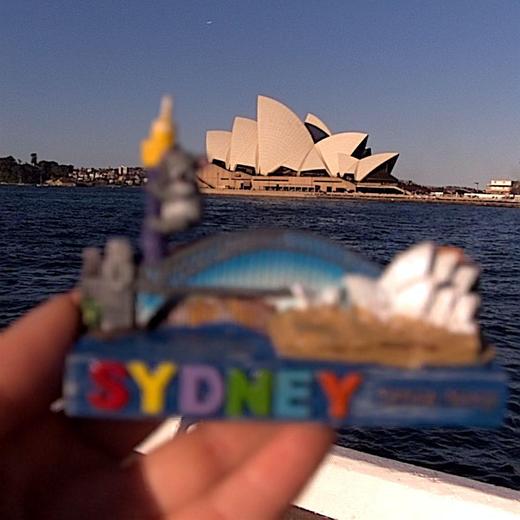}}%
  \subfigure[Fusion image]{\label{fig:subfig:13}
    \includegraphics[width=0.32\linewidth]{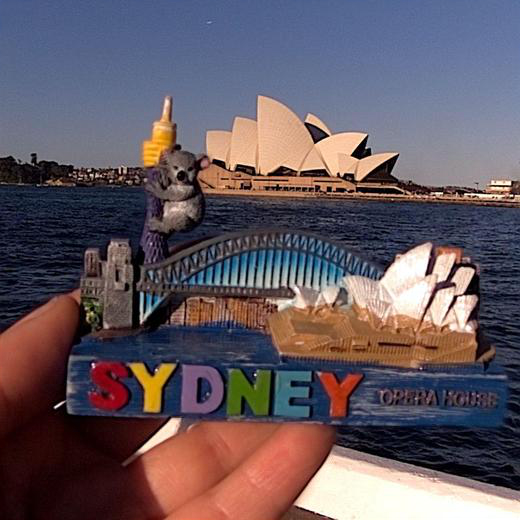}}
  \caption{An example of multi-focus image fusion with two source images; the fusion image was obtained by our proposed MMF-Net.}
\label{fig:1} 
\end{figure}

\IEEEPARstart{W}{hen} photos are taken with cameras, all-in-focus images are often desired as the output, in particular for a large number of computer vision tasks, such as localization, detection and segmentation \cite{hassen2015objective}. However, it is usually hard to obtain an all-in-focus images from a single camera, since the depth of field of the camera is limited \cite{bouzos2019conditional}. Multi-focus image fusion is the approach to generating an all-in-focus image from several images taken on the same scene but focused at different depths, as shown in Fig.~\ref{fig:1} via an example of the fusion image obtained from two source images. 

Existing multi-focus image fusion (MFIF) methods can be broadly categorized into three groups, i.e., transform domain algorithms, spatial domain algorithms, and convolutional neural network (CNN) based algorithms \cite{kaur2015comparative}.

The transform domain MFIF algorithms first decompose the source images and then fuse the results according to some handcrafted features. Typical transform domain MFIF algorithms include the non-subsampled contourlet transform (NSCT) method \cite{ZHANG20091334}, the sparse representation (SR) method \cite{RN35, zhang2016robust} and the combined NSCT-SR method \cite{LIU2015147}. Due to the imperfection of transformations and handcrafted features, these algorithms often produce non-realistic results, even in the areas far away from the focused/defocused boundary (FDB).

The spatial domain MFIF algorithms include region based methods and pixel based methods. Region based MFIF algorithms suffer from the blocking effect \cite{RN85}. Pixel based MFIF algorithms first get a 0/1 discrete decision map for fusion \cite{RN43}, and then fuse the source images. The guided filtering (GF) method \cite{RN80} and the dense SIFT (DSIFT) method \cite{RN43} are typical pixel based algorithms. Compared with the transform domain algorithms, the fusion results from the pixel based algorithms are usually better, However, due to the defocus effect, none of the source images are clear in the areas near the FDB, and consequently, the fusion results of these methods are often unclear in these areas.

The CNN is first explored to extract defocus descriptors in a data-driven way in \cite{wang2010multi}. Existing neural network based MFIF algorithms can be divided into two groups: decision map based algorithms and end-to-end algorithms. Decision map based algorithms \cite{RN91, RN497, 8625482} produce the decision map first as with the pixel based algorithms, which, also similarly to the pixel-based algorithms, lead to the unclear FDB. End-to-end algorithms \cite{ma2019sesf, yan2018unsupervised, zhao2018multi} directly obtain the fusion results, but the results are unfortunately not realistic, as with the transfer domain algorithms. Moreover, since it is hard to get a large number of all-in-focus ground truth images for training, data generation methods need to be adopted in these CNN based algorithms \cite{RN91, RN497, 8625482, ma2019sesf, yan2018unsupervised, zhao2018multi}; however, these methods do not imitate the complex situation of defocus spread near the FDB, and thus some unnatural and unrealistic training data limits the performance of these networks.

In this paper, to address the issue of unsatisfied fusion results near the FDB, we first present a discussion about the difficulties around the FDB. Then an $\alpha$-matte boundary defocus model is proposed to simulate the defocus spread effect near the FDB. Based on the $\alpha$-matte defocus model and the generated training data, we develop a cascaded network for MFIF, which is called the Matte Model Fusion Net (MMF-Net). The technical underpinnings and contributions of our work are two-fold:

Firstly, a novel $\alpha$-matte boundary defocus spread model is proposed. Compared with existing defocus models, the proposed $\alpha$-matte model is the first one to specifically model the difference in defocus spread when the defocus happens to the foreground or the background. Therefore, the $\alpha$-matte model can generate simulated defocus images with the valid defocus spread near the FDB, which can be then used as training data to train deep neural networks. 

Secondly, based on the proposed $\alpha$-matte defocus model and the generated training data, a cascaded boundary aware convolutional network termed MMF-Net is designed and trained, to obtain clear fusion results in areas both far away from and near the FDB. Compared with the existing end-to-end CNN algorithms, the proposed MMF-Net generates a guidance map first, acquiring clear and realistic fusion results in areas far away from the FDB. Compared with the decision map based CNN algorithms, the MMF-Net generates the areas near the FDB directly from the source images, like the end to end CNN algorithms, and thus achieve clearer and reasonable fusion results than any source images near the FDB. 

Experiments demonstrate that, on the benchmark dataset, the proposed MMF-Net outperforms the state-of-the-art methods, both qualitatively and quantitatively.

The rest of this paper is organised as follows. In section~\ref{s:model}, we present the issue to be addressed in this paper, introduce the proposed $\alpha$-matte boundary defocus model, and conduct comparison with the existing defocus models on the simulated scene. In section~\ref{s:matte}, we present the proposed MMF-Net and discuss the corresponding loss functions. Experimental studies including the method comparison are conducted in section~\ref{s:exp}, and conclusions and future work are drawn in section~\ref{s:conclusions}.

\section{$\alpha$-Matte Boundary Defocus Model}\label{s:model}

In this section, we first discuss the defocus spread effect around the focused/defocused boundary (FDB), and why it is difficult to deal with. Then we briefly introduce and analyze two existing defocus models: the one-parameter defocus model \cite{zhuo2011defocus} and our previous two-parameter defocus model \cite{Liu-TIP-2016}. Finally, a novel $\alpha$-matte boundary defocus spread model is proposed based on the above discussion and analysis. Simulation experiments on a well-designed scene shows that the proposed $\alpha$-matte model could generate a defocus effect near the FDB much more realistic than the existing defocus models.

\subsection{The Focus/Defocus Boundary (FDB)}

Existing CNN methods cannot obtain realistic and clear fusion results, in particularly for the areas near the FDB. There are three main reasons for this issue. 

\begin{figure}[htbp]
\centering 
  \subfigure[Foreground focus]{\label{fig:subfig:21}
    \includegraphics[width=0.48\linewidth]{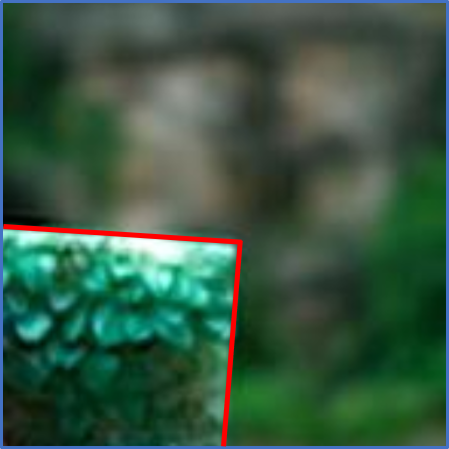}}%
  \subfigure[Background focus]{\label{fig:subfig:22}
    \includegraphics[width=0.48\linewidth]{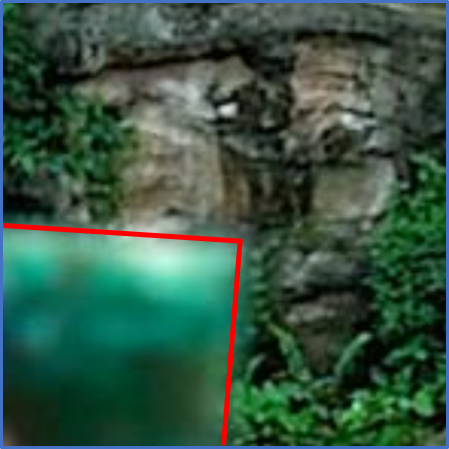}}%
  \caption{Different defocus spread effects when the foreground or the background is out of focus, shown in enlarged real-world images. The in-focus boundary of foreground object is labeled red.}
\label{fig:2} 
\end{figure}

Firstly, the situations are quite different between patches far away from and near the FDB, and it is unwise to deal with an area near the FDB and an area far away from the FDB together, as stated in our previous work \cite{MA2019125}. For the patches far away from the FDB, the patches are totally focused or defocused. Consequently, the defocus of the patch is homogeneous. On the contrary, for patches near the FDB, both focused area and defocused area exist. Therefore, it is hard to separate the focus area and the defocus area at the pixel level.

Secondly, there is a blurry area along the FDB, which is unclear in both source images A and B, because the defocus effect will spread. When the foreground object is in focus, but the background is defocused, the foreground object will not be influenced by the defocus of the background, as shown in Fig.~\ref{fig:subfig:21}. In contrast, when the foreground object is defocused, the defocus spread effect will lead to a blurry object bigger than the original focused object, as shown in Fig.~\ref{fig:subfig:22}, in which we highlight the boundary of foreground objects in red to show the difference in defocus spread. As a result, there is an area blurry in both source images A and B, along the outside of the foreground objects, one due to the defocus of the background and the other due to the defocus spread from the defocused foreground objects.

Thirdly, when the foreground object is out of focus (Fig.~\ref{fig:subfig:22}), it will be influenced slightly by the background on the inside of the original boundary too, compared with Fig.~\ref{fig:subfig:21}. That is because the moving of the camera lens when changing the focal length will lead to a small change of scene. As a result, although the scene varies very slightly, there will be a mismatch for areas around the FDB when a fusion is conducted on the source images.

The defocus spread effect of foreground objects makes it hard to obtain a clear result near the FDB. Some of the existing methods \cite{RN43, RN91, RN497} choose the pixel directly from one of the source images, and thus the fusion results near the FDB will be blurry with artifact. Some post processing methods \cite{RN497}, such as guided filter \cite{he2010guided}, are to derive smooth but still unclear FDB. Even using a weighted average of the source images as the fusion result \cite{LIU2015147, RN80, zhao2018multi}, the blur will still remain. To address this issue, we need to carefully model the defocus spread for the areas near the FDB, in order to generate a large number of realistic training images to train the neural networks for the MFIF. The performance of these data-driven methods is highly dependent on the training datasets, therefore, how closely the model can simulate the reality is of vital importance.

There are several defocus spread models based on which the training datasets for CNN-based MFIF methods are generated. Existing models include the one-parameter defocus model and the two-parameter defocus model. Usually, these models are employed to generate training dataset for MFIF methods. However, these models are not always valid, especially for the areas near the FDB.

\subsection{One-parameter Defocus Model}

The typical one-parameter 2D linear space-invariant defocus model \cite{zhuo2011defocus} can be characterized with a space-invariant Point Spread Function (PSF) \cite{chen2017robust}:
\begin{eqnarray}
\label{G1D}
    I(x, y) = h(x,y) \otimes f(x, y) + n(x, y),
\end{eqnarray}
where $I(x,y)$ is the defocused image at pixel $(x, y)$, $f(x,y)$ is the original image without the defocus effect, $n(x,y)$ is the additive noise, $h(x,y)$ is the defocus kernel, and $\otimes$ denotes the convolution operator. In practice, the defocus kernel $h(x,y)$ is usually approximated with a 2D isotropic Gaussian kernel
\begin{eqnarray}
    h(x,y) = G(x, y; \sigma) = \frac{1}{2\pi\sigma^2}\exp(-\frac{x^2+y^2}{2\sigma^2}),
\end{eqnarray}
where $\sigma$ is the standard deviation, which describes the defocus amount and is related to the distance between the object and the camera.

Several CNN-based MFIF methods such as \cite{RN497, RN91} employed this model to generate training data. In \cite{RN497}, the original images from the ImageNet dataset \cite{deng2009imagenet} are directly reblurred with a random Gaussian kernel for the defocus effect. This is a simple way to generate data, but the relation between the defocus amount and the depth is not considered. \cite{RN91} noticed that usually the defocus did not happen over the whole image, and thus they reblurred the input image only in a predetermined area. However, the boundary of the predetermined area usually do not coincide with the real boundary of the object in the image.

\subsection{Two-parameter Defocus Model}

Since the defocus level is related to the depth between the objects and the camera, different objects can have different defocus levels. In our previous work on defocus map estimation \cite{Liu-TIP-2016}, a two-parameter defocus model was proposed to describe the defocus spread effect for the area near the object boundaries, with a PSF with two parameters used to describe the different defocus levels on the two sides of a boundary. For an ideal 2D boundary,
\begin{eqnarray}
\begin{aligned}
    f(x, y) = &f_A(x, y) u(ax+by+c) +\\
    &f_B(x, y) u(-ax-by-c),
\end{aligned}
\end{eqnarray}
and the defocused boundary will be
\begin{eqnarray}
\label{G2D}
\begin{aligned}
    I(x, y) = & f_A(x, y) u(ax+by+c) \otimes h_A(x,y)+\\
    &f_B(x, y) u(-ax-by-c) \otimes h_B(x,y),
\end{aligned}
\end{eqnarray}
where $u(\cdot)$ is the step function, $ax+by+c=0$ is the line corresponding to the boundary, and $f_A(x,y)$ and $f_B(x,y)$ are the original image areas at the different sides of the boundary, respectively. In \cite{Liu-TIP-2016}, the defocus kernels $h_A(x,y)$ and $h_B(x,y)$ are approximated with two different 2D isotropic Gaussian kernels:
\begin{eqnarray}
    h_A(x,y) =  G_A(x, y; \sigma_A) = \frac{1}{2\pi\sigma_A^2}\exp(-\frac{x^2+y^2}{2\sigma_A^2}),\\
    h_B(x,y) =  G_B(x, y; \sigma_B) = \frac{1}{2\pi\sigma_B^2}\exp(-\frac{x^2+y^2}{2\sigma_B^2}).
\end{eqnarray}

Taking advantage of this model, we generate the training data for MFIF in our previous work \cite{MA2019125}, in which the foreground objects or the background are first blurred with the Gaussian function and then spliced together. Consequently, the defocus level only changes alongside object boundaries, which is closer to the reality than \cite{RN497, RN91}. However, this two-parameter model cannot describe the difference between out-of-focus foreground (with in-focus background) and out-of-focus background (with in-focus foreground).

\subsection{An $\alpha$-Matte Boundary Defocus Spread Model}

As we have discussed above, existing defocus models focus on the intensity at each pixel, rather than the defocus spread along the boundary of the foreground objects. To simulate the defocus effect, a new model should focus on several specific issues: the defocus spread across the FDB, the blurry area along the FDB, and the different spread situations of the FDB when defocus happens to the foreground or the background. To address these issues, we first propose a novel $\alpha$-matte boundary defocus spread model to simulate the defocus process. 

In the proposed $\alpha$-matte model, we assume that there is a transmission matte $\alpha_n$ for every surface $S_n$ parallel to the focal plane, where $n(=1,\ldots,N)$ is the order of the surface. Firstly, we assume that when a surface is in focus, for the surface without object on it, the matte values is zero, and for the surface with objects that no light can go through, the matte on the object pixel is one. Secondly, the defocus kernel $h_n(x,y)$ for a defocused surface $S_{n}(x,y)$ is a 2D isotropic Gaussian kernel $G(x, y; \sigma_{n})$:
\begin{eqnarray}
    h_n(x,y) = G(x, y; \sigma_n) = \frac{1}{2\pi\sigma_n^2}\exp(-\frac{x^2+y^2}{2\sigma_n^2}).
\end{eqnarray}
Thirdly, the defocus effects are the same for the RGB surface $S_{n}$ and the matte $\alpha_{n}$:
\begin{eqnarray}
    \begin{aligned}
        \alpha^0_{n}(x, y) &= h_n(x,y) \otimes \alpha_{n}^{c}(x, y) \\
        &= G(x,y;\sigma_{n}) \otimes \alpha_{n}^{c}(x, y),
    \end{aligned}\\
    \begin{aligned}
        S_{n}(x, y) &= h_n(x,y)\otimes S_{n}^{c}(x, y) \\
        &= G(x,y;\sigma_{n}) \otimes S_{n}^{c}(x, y),
    \end{aligned}
\end{eqnarray}
where $\alpha^{c}_n$ is the clear matte on the in-focus clear surface $S^{c}_n$, and $\alpha_{n}^{0}$ denotes the matte before considering any defocus spread effects from the objects in front of $S_n$).

As have been shown in Fig.~\ref{fig:2}, the out-of-focus objects in front can affect objects behind, but the out-of-focus objects behind cannot affect objects in front. Therefore, we make the clear RGB surface $S^{c}_n$ and the clear matte $\alpha_{n}^{c}$ out of focus one by one from near to far. The final matte $\alpha_n$ will be an aggregation of all effects from those mattes in front, that is, the intersection between its own $\alpha_n^{0}$ and the complementary set for the summation of all the mattes in front:
\begin{eqnarray}
 \alpha_{n} = \alpha_{n}^{0} (1-\sum_{t=0}^{n-1} \alpha_{t}),\ \ {\rm with}\ \alpha_0=0,\ n=1,\ldots, N. 
\end{eqnarray}
This means that the defocus of matte would only influence the mattes behind it, as with the reality. Therefore, the captured image $I$ at each pixel in the photo will be a summation of the pixel values on all surfaces $S_{n}$:
\begin{equation}
\label{I}
 I = \sum_{n=1}^{N} I_n = \sum_{n=1}^{N}[(1-\sum_{t=0}^{n-1} \alpha_{t}) S_n]. 
\end{equation}
Particularly, image $I$ with only two valid surfaces (foreground surface $S_{FG}$ and background surface $S_{BG}$) is
\begin{equation}
\label{equation2}
I = S_{FG} + (1 - \alpha_{FG}) S_{BG}, 
\end{equation}
where $\alpha_{FG}$ is the matte of the foreground, and this is the model we used for data generation.

\subsection{Comparison of Defocus Models}

\begin{figure}[htbp]
\centering 
  \subfigure[The stimulated scene]{\label{fig:subfig:71}
    \includegraphics[width=0.92\linewidth]{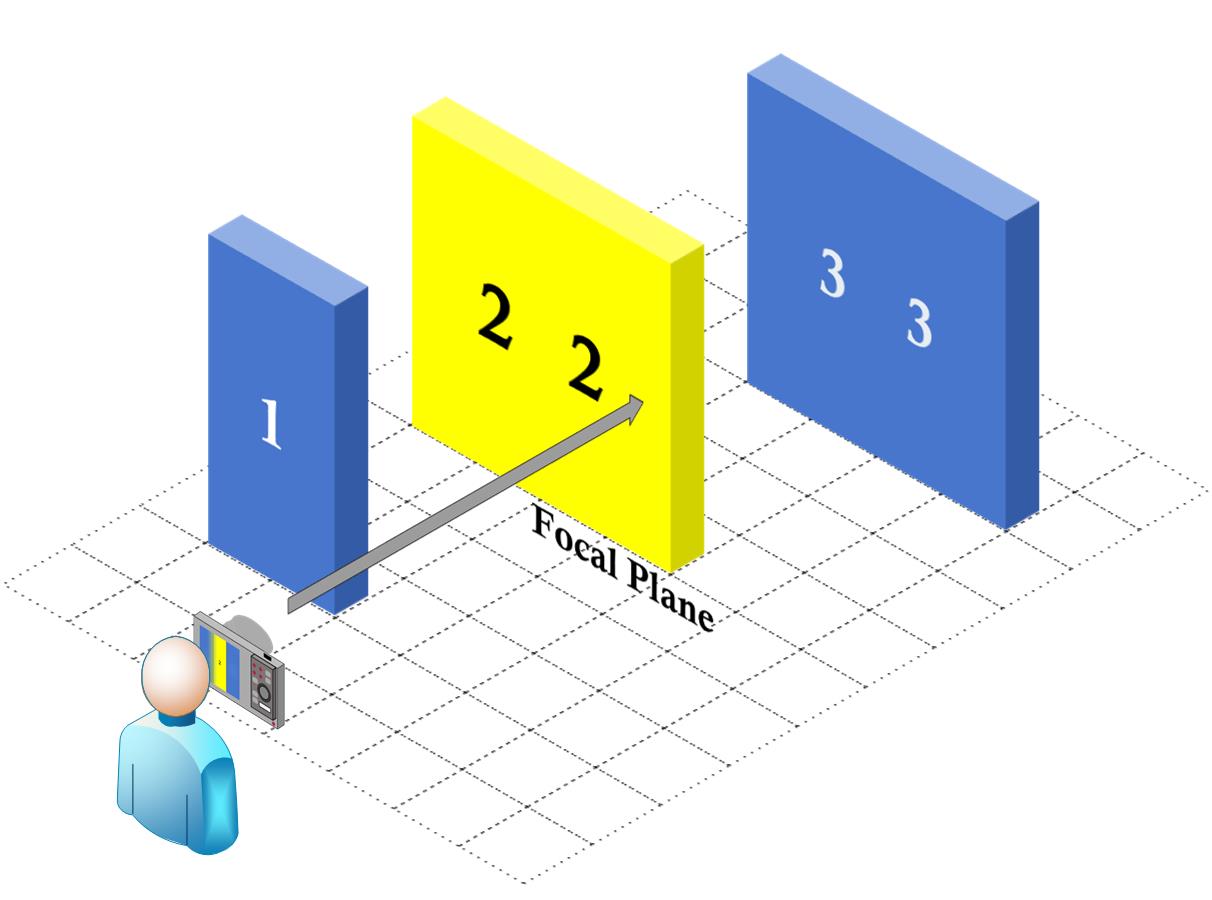}}\\
  \subfigure[All-in-focus image]{\label{fig:subfig:72}
    \includegraphics[width=0.48\linewidth]{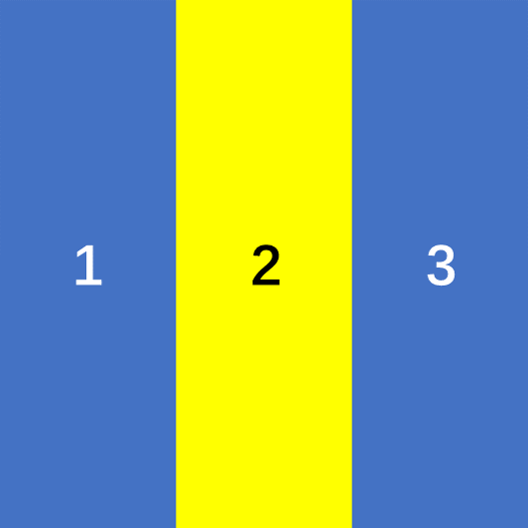}}
  \subfigure[One-parameter model]{\label{fig:subfig:74}
    \includegraphics[width=0.48\linewidth]{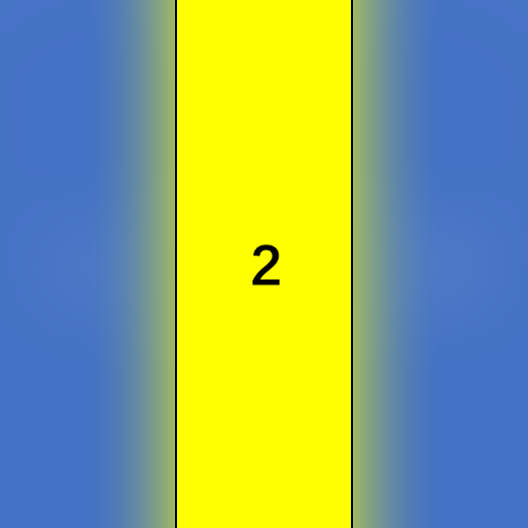}}\\
  \subfigure[Two-parameter model]{\label{fig:subfig:75}
    \includegraphics[width=0.48\linewidth]{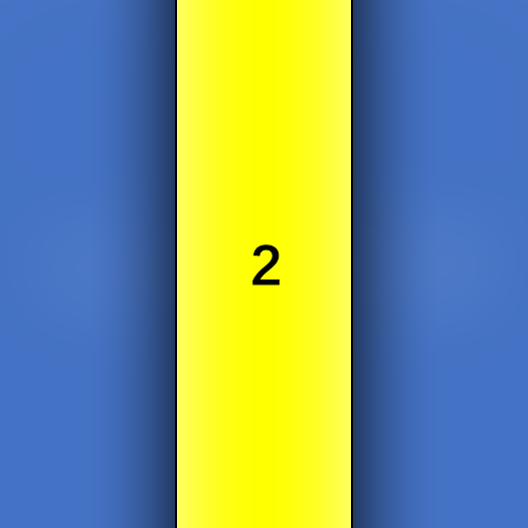}}
  \subfigure[$\alpha$-matte defocus model]{\label{fig:subfig:73}
    \includegraphics[width=0.48\linewidth]{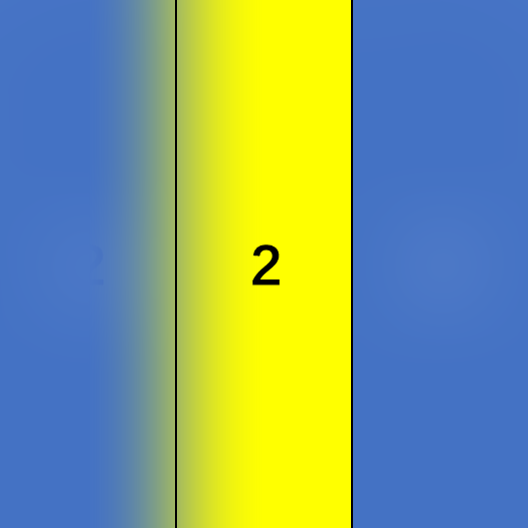}}
  \caption{Comparison of defocus models in the stimulated scene with camera focusing on object 2. The boundary between object 1 and object 2 is different from that between object 2 and object 3. The proposed $\alpha$-matte model precisely simulates the defocus spread effect across the boundary between object 1 and object 2 (foreground out of focus), as well as the clear boundary between object 2 and object 3 (foreground in focus). The black lines are added to show the original boundary of clear objects.}
\label{fig:7} 
\end{figure}

To show the difference between the proposed $\alpha$-matte defocus spread model and the existing defocus models, we stimulate it with a simple scene. In the stimulated scene, there are three objects, as shown in Fig.~\ref{fig:subfig:71}. Object 1 is close to the camera, object 3 is far away from the camera, and object 2 is set between object 1 and object 3. The all-in-focus image will be Fig.~\ref{fig:subfig:72}, which cannot be taken by the camera with limited focal of length directly. We focus the camera on the object 2, so the defocus spread effect near the FDB varies. For the boundary between object 1 and object 2, the foreground is out of focus and the background is in focus. In the contrary, for the boundary between object 2 and object 3, the foreground is in focus and the background is out of focus. 

According to the one-parameter defocus model, the simulated image will be as Fig.~\ref{fig:subfig:74}. The model could simulate the defocus effect in object 1 and object 3, which are out of focus. But the one-parameter defocus model cannot show the defocus spread effect at all: in the area near the FDB boundary in object 2, there is no influence of defocus at all. Moreover, this model has no concern about whether the defocus object is in front or behind of an in-focus object, with the FDB all the same as shown in Fig.~\ref{fig:subfig:74}.

According to the two-parameter defocus model, the simulated image will be as Fig.~\ref{fig:subfig:75}. The defocus spread effect could be simulated at both sides of the FDB, but this defocus model suffers from the anti-gradient effect. Especially for the situation shown in this simulated scene, object 2 is in focus and shows no influence on object 1; and the defocus spread of object 1 is unclear since object 2 is in focus. Also, the two-parameter defocus model also has no concern about whether the defocus object is in front or behind, with the FDB all the same in Fig.~\ref{fig:subfig:75}.

\begin{figure*}[htbp]
\centering 
  \subfigure[$S^c_n$]{\label{fig:subfig:82}
    \includegraphics[width=0.152\linewidth]{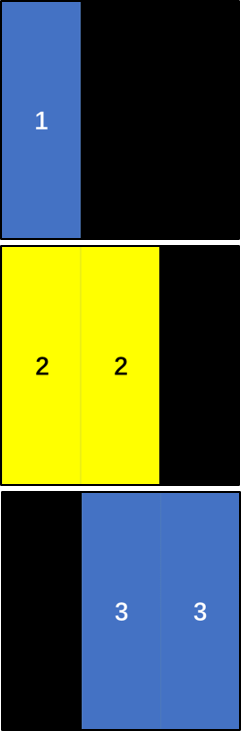}}
  \subfigure[$\alpha^c_n$]{\label{fig:subfig:81}
    \includegraphics[width=0.152\linewidth]{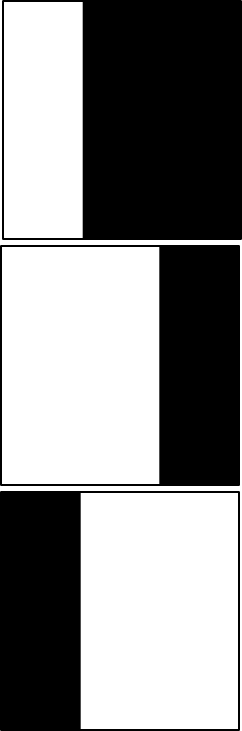}}
  \subfigure[$S_n$]{\label{fig:subfig:84}
    \includegraphics[width=0.152\linewidth]{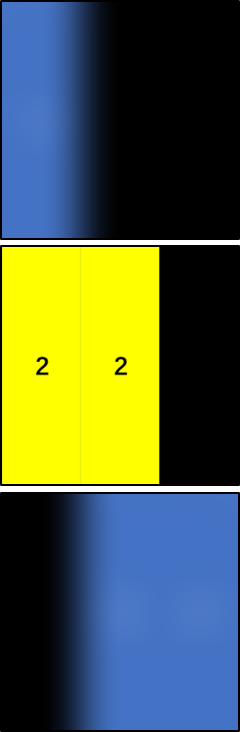}}
  \subfigure[$\alpha^0_n$]{\label{fig:subfig:83}
    \includegraphics[width=0.152\linewidth]{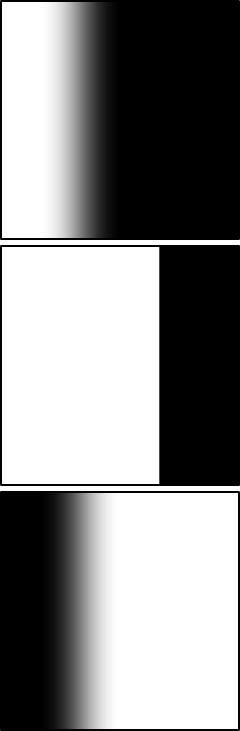}}
  \subfigure[$\alpha_n$]{\label{fig:subfig:85}
    \includegraphics[width=0.152\linewidth]{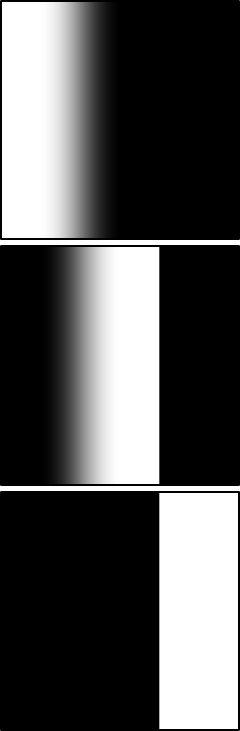}}
  \subfigure[$I_n$]{\label{fig:subfig:86}
    \includegraphics[width=0.152\linewidth]{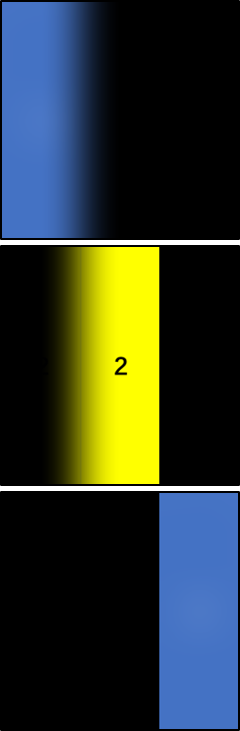}}
  \caption{Imaging process of the proposed $\alpha$-matte defocus model in a stimulated scene with three surfaces from top down. The photo $I$ (Fig.~\ref{fig:subfig:73}) is the summation of the last column $I_n$.}
\label{fig:8} 
\end{figure*}

The defocus simulated image of the proposed $\alpha$-matte defocus spread model is shown in Fig.~\ref{fig:subfig:73}, from which we can observe two patterns.

Firstly, for the FDB between object 1 and object 2, the defocus spread effect is at both sides: the effect in object 1 is due to the defocused object 1 and the moving camera lens; and the effect in object 2 is due to the spread effect of the defocused object 1. That is, because  object 1 (foreground object) is out of focus and object 2 (background object) is in focus, the defocus will spread to the area of object 2 near the FDB, and the yellow color of object 2 will also have a slight influence on the object 1. In other words, if  object 2 is with texture, we will notice that the defocus spread to object 1 is in fact the texture behind the object 1 compared with the all-in-focus image; and the proposed $\alpha$-matte model could also simulate the scene change when the focal length changes.

Secondly, for the FDB between the in-focus object 2 and the defocused object 3, the defocus spread has no effect on either side. The object 2 (foreground object) is in focus and object 3 (background object) is out of focus. Consequently, the defocus will not spread to the area of object 3 near the FDB, as object 2 is in focus; and object 3 cannot influence the object 2 too, as object 3, although out of focus, is behind the in-focus object 2. 

In short, the proposed defocus model simulates very well the difference of defocus spread when defocus is happened to the foreground or the background.

Using the same simulated scene, we also show in Fig.~\ref{fig:8} the image composition process with the proposed $\alpha$-matte defocus spread model. For every surface $S_n$ parallel to the focal plane, the before-defocus object surfaces $S_n^c$ is shown in Fig.~\ref{fig:subfig:82}, and the before-defocus matte $\alpha_n^c$ is shown in Fig.~\ref{fig:subfig:81}. 

Then the camera is set to focus at object 2, and at the same time object 1 and object 3 are out of focus. As we have mentioned above, a 2D isotropic Gaussian kernel $G(x, y; \sigma_{n})$ is applied to the clear surface $S_n^c$ and the clear matte $\alpha_n^c$ at the same time. The defocus exist when $n=1$ or $n=3$, and object 2 is in focus. The surfaces are shown in Fig.~\ref{fig:subfig:84}, and the mattes are shown in Fig.~\ref{fig:subfig:83}.

To compose the final image taken by the simulated camera, we first get the $\alpha_n$ which is the intersection between $\alpha_n^{0}$ and the complementary set for the summation of all the mattes in front. The calculation is done one by one from the surface in the front. After that, the image are generated using Equation (\ref{I}). The compositions of each surface are shown in Fig.~\ref{fig:subfig:86}. And the final defocus image is the summation of $S_1$, $S_2$ and $S_3$, as we have shown in Fig.~\ref{fig:subfig:73}.

\begin{figure*}[htbp]
\centering 
   \includegraphics[width=0.96\linewidth]{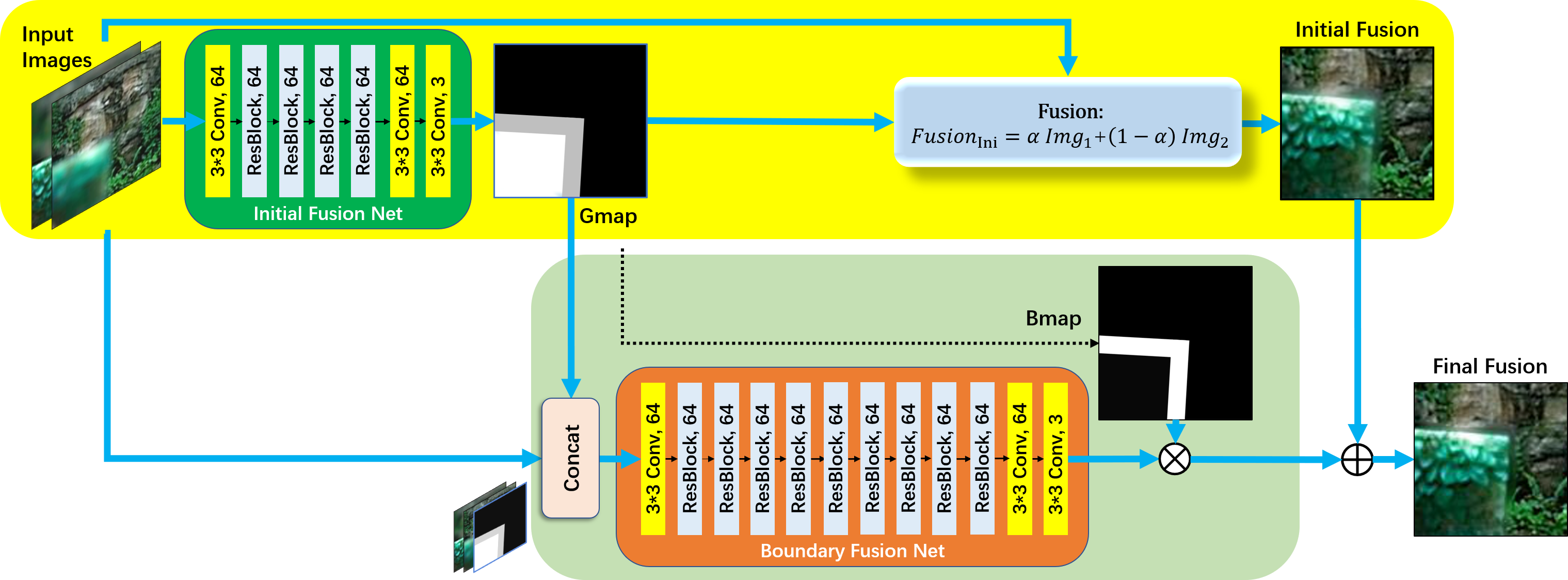}
   \caption{The block diagram of the proposed cascaded boundary aware multi-focus fusion network MMF-Net. The initial fusion sub-net is to derive the guidance map $Gmap$, and then the boundary fusion sub-net is to help refine the fusion result near the FDB.}
\label{fig:3}
\end{figure*}

\section{MMF-Net: the Matte Model Based Cascaded Fusion Net}\label{s:matte}

Based on the proposed defocus model and its generated training data, the cascaded convolutional fusion net (MMF-Net) is developed and trained. In this section, the structure of the proposed MMF-Net is firstly introduced, and then the loss functions used to train the MMF-Net are discussed. Our MMF-Net aims to achieve realistic and clear results in both the areas near and far away from the FDB. 

\subsection{The Network Structure}

The structure of the proposed MMF-Net is shown in Fig.~\ref{fig:3}. 

Two generated source images are firstly input into an Initial Fusion sub-net aiming to generate a guidance map $Gmap$. In the guidance map, the pixel-wised value would be 1 if source image A is focused and source image B is defocused, and be 0 if source image A is defocused and source image B is focused, and the area near the FDB is given an $\alpha$ values of 0.5. Then we use this guidance map to generate a initial fusion result:
\begin{equation}
Fusion_{Ini} = Gmap\times  I_1 + (1 - Gmap)\times  I_2,
\end{equation}
where $I_1$ and $I_2$ are the two source images.

Then the two source images are concatenated with the guidance map as the input of the Boundary Fusion sub-net. The output of the Boundary Fusion sub-net is masked and weighted by the boundary map $Bmap$, and then added to the initial fusion result. The boundary map is calculated with the guidance map as
\begin{equation}
    Bmap = 1-|2\times Gmap-1|.
\end{equation}
In this way, no matter it is focused in which source image, the value will be 0 for the areas far away from the FDB. Hence, only the pixels of boundary areas in the initial fusion result will be revised by the output of the Boundary Fusion sub-net. That is, for the areas far away from the FDB, the final fusion results $Fusion_{Fin}$ of MMF-Net will be completely decided by the focused part of the source images. In the meantime, for the areas near the FDB, the final fusion results $Fusion_{Fin}$ are obtained through enhancing the initial fusion results by the output of the Boundary Fusion sub-net, which was hard for existing methods as they do not specifically treat the FDB.

In our implementation, typical residual blocks \cite{RN618} are employed.
The Initial Fusion sub-net contains 1 convolutional layers, 4 residual blocks, and 2 convolutional layers; the Boundary Fusion sub-net contains 1 convolutional layers, 12 residual blocks, and 2 convolutional layers. The kernel size of every residual block is 64.

\subsection{Loss Functions}

The loss function that we use in the training process contains three components: the matte loss $Loss_{matte}$, the initial fusion loss $Loss_{Ini}$, and the weighted final fusion loss $Loss_{W}$:
\begin{equation}
    Loss = \lambda_{1} \times Loss_{matte} + \lambda_{2} \times Loss_{Ini} + Loss_{W},
\end{equation}
where $\lambda_1$ and $\lambda_2$ are trade-off parameters.

Firstly, for the Initial Fusion sub-net, matte loss $Loss_{matte}$ and initial fusion loss $Loss_{Ini}$ are used:
\begin{eqnarray}
    Loss_{matte} &=& mean(|matte_{Ini}-matte_{GT}|),\\
    Loss_{Ini} &=& mean((Fusion_{Ini}-Fusion_{GT})^2).
\end{eqnarray}
We choose the L1 norm for matte as it is a discrete value in the ground truth, and choose the L2 norm for the fusion results as usual.

Secondly, in order to supervise the final fusion result more precisely, we use a weighted fusion loss $Loss_{W}$. The area near the FDB is much more difficult to fused than the other areas, so its weight $W$ should be larger than those for the areas far away from the FDB:
\begin{equation}
    Loss_{W} = W \times mean((Fusion_{Fin}-Fusion_{GT})^2),
\end{equation}
where the weight $W$ is simply calculated as follows:
\begin{equation}
    W = \frac{1+(k-1)\times (1-|2\times matte_{Ini}-1|)}{k},
\end{equation}
where $k$ is the weight contrast parameter of the FDB area. The max weight for the area near the FDB will be close to 1, as the matte values will be close to 0.5; and for the area far away from the FDB, the weight will be close to $1/k$, as the matte values will be close to either 1 or 0. 

When training the fusion model, we use $\lambda_{1}=\lambda_{2}=0.2$ and $k=5$; that is, more attention is given to the FDB area because it is difficult to obtain.

\section{Experimental Studies}\label{s:exp}

This section will present the dataset generation, implementation settings, comparison settings and experimental results.


\subsection{Dataset Generation}

A good training dataset should represent comprehensive situations of the task. The best choice of training data is real photos. However there are few multi-focus source images for fusion, and the ground truth needs to be labeled manually, which is very costly. Therefore, a feasible way is to generate artificial training images that are similar to the reality yet easy to obtain. In our case, a dataset of foreground images with ground truth is used, and some images without obvious defocus are chosen as the background dataset. Both the original foreground ($FG^{C}$) and the background ($BG^{C}$) images are first processed by Gaussian filters with kernel $G(x, y; \sigma)$ for the blurred images:
\begin{eqnarray}
FG^{B} (x, y) &=& G(x, y; \sigma) \otimes FG^{C} (x, y), \\
BG^{B} (x, y) &=& G(x, y; \sigma) \otimes BG^{C} (x, y).
\end{eqnarray}

\begin{figure*}[htbp]
\centering 
  \subfigure[$ImgS_1$]{\label{fig:subfig:91}
    \includegraphics[width=0.130\linewidth]{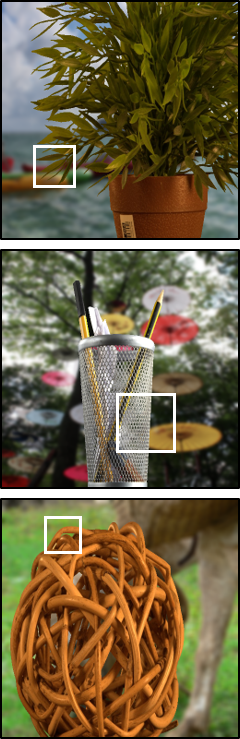}}
  \subfigure[Enlarged $ImgS_1$]{\label{fig:subfig:911}
    \includegraphics[width=0.130\linewidth]{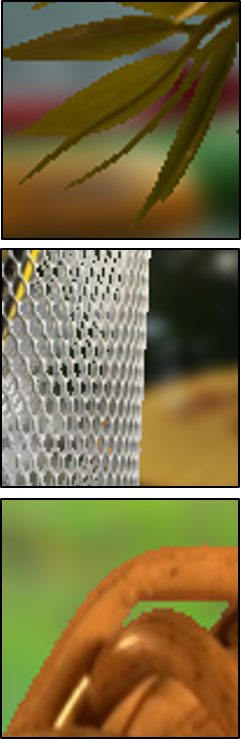}}
  \subfigure[$ImgS_2$]{\label{fig:subfig:92}
    \includegraphics[width=0.130\linewidth]{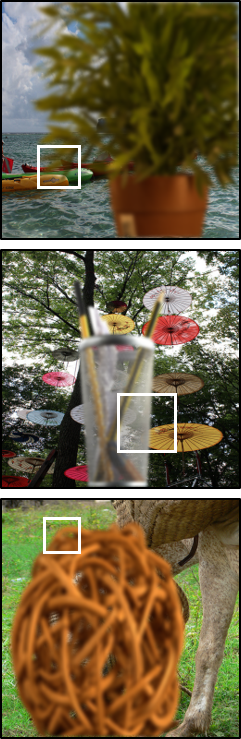}}
  \subfigure[Enlarged $ImgS_2$]{\label{fig:subfig:922}
    \includegraphics[width=0.130\linewidth]{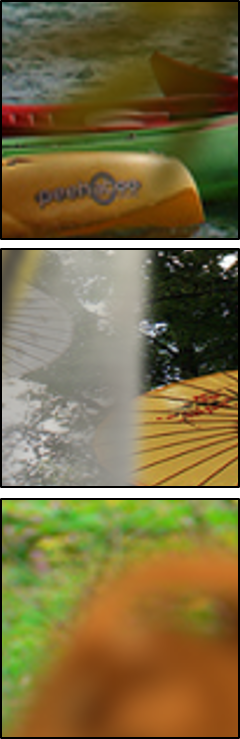}}
  \subfigure[$G_{map}$]{\label{fig:subfig:93}
    \includegraphics[width=0.130\linewidth]{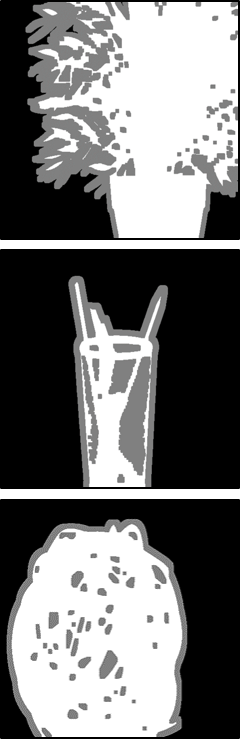}}
  \subfigure[$G'_{map}$]{\label{fig:subfig:94}
    \includegraphics[width=0.130\linewidth]{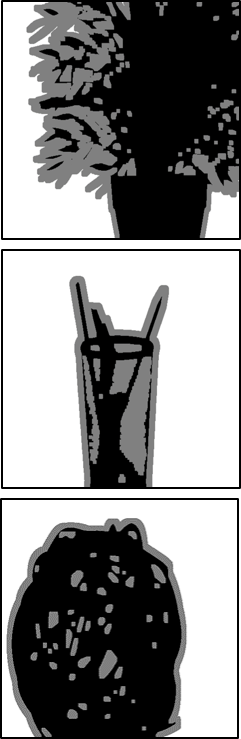}}
  \subfigure[Fusion GT]{\label{fig:subfig:95}
    \includegraphics[width=0.130\linewidth]{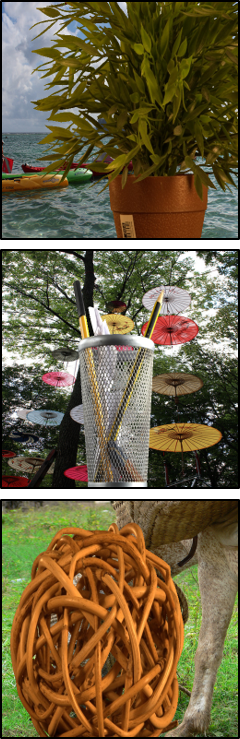}}
  \caption{Examples from the generated training dataset. The defocus spread effect is slight but can be seen in the enlarged images.}
\label{fig:9} 
\end{figure*}

When the foreground is in focus, the ground truth is the same as the matte $\alpha^{C}$; and when the background is in focus, the matte $\alpha^{B}$ is the blurred ground truth with the Gaussian kernel $G(x, y; \sigma)$:
\begin{eqnarray}
\alpha^{B} (x, y) &=& G(x, y; \sigma) \otimes \alpha^{C} (x, y).
\end{eqnarray}

Then the source images are generated using Equation (\ref{equation2}) according to the matte ($\alpha^{C}$ or $\alpha^{B}$) pixel by pixel. Source image 1 ($ImgS_1$) is with the in-focus foreground and the out-of-focus background; and source image 2 ($ImgS_1$) is with the out-of-focus foreground and the in-focus background:
\begin{eqnarray}
ImgS_1 = FG^{C} + (1 - \alpha^{C}) BG^{B}, \\
ImgS_2 = FG^{B} + (1 - \alpha^{B}) BG^{C}.
\end{eqnarray}
Examples of the source image pairs of the generated training dataset are also shown in Fig.~\ref{fig:subfig:91} and Fig.~\ref{fig:subfig:92}. The defocus spread effect when the foreground is out of focus can be seen in the enlarged images in Fig.~\ref{fig:subfig:922}, compared with the one when the foreground is in focus as shown in Fig.~\ref{fig:subfig:911}. Fusion ground truth (Fig.~\ref{fig:subfig:95}) is generate with the matte $\alpha^{C}$:
\begin{eqnarray}
GT = FG^{C} + (1 - \alpha^{C}) BG^{C}.
\end{eqnarray}

The guidance maps (Fig.~\ref{fig:subfig:93} and Fig.~\ref{fig:subfig:94}) are created at the same time. In the blurred matte $\alpha^{B}$, the value in (0, 1) are set to 0.5 as the guidance map. If the foreground is in focus in source image A and out of focus in source image B, the guidance map $Gmap$ will be
\begin{equation}
Gmap(x,y)=
\left\{
    \begin{array}{lr}
    0,   &   \alpha^B(x,y) = 0\\
    0.5, &   0 < \alpha^B(x,y) < 1\\
    1,   &   \alpha^B(x,y) = 1
    \end{array}
\right.
.
\end{equation}
On the other hand, if the foreground is defocused in source image A and focused in source image B, the guidance map $Gmap'$ will be the opposite of $Gmap$:
\begin{equation}
Gmap'(x,y)=
\left\{
    \begin{array}{lr}
    1,   &   \alpha^B(x,y) = 0\\
    0.5, &   0 < \alpha^B(x,y) < 1\\
    0,   &   \alpha^B(x,y) = 1
    \end{array}
\right.
.
\end{equation}

\begin{figure*}[htbp]
\centering 
  \subfigure[Source A]{\label{fig:subfig:41}
    \includegraphics[width=0.19\linewidth]{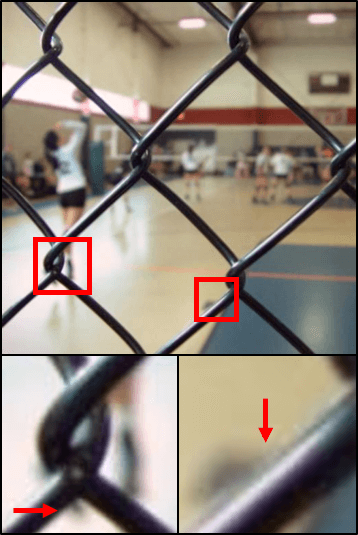}}%
  \subfigure[Source B]{\label{fig:subfig:42}
    \includegraphics[width=0.19\linewidth]{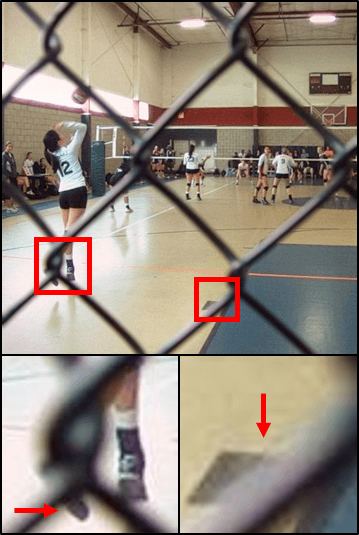}}
  \subfigure[NSCT]{\label{fig:subfig:43}
    \includegraphics[width=0.19\linewidth]{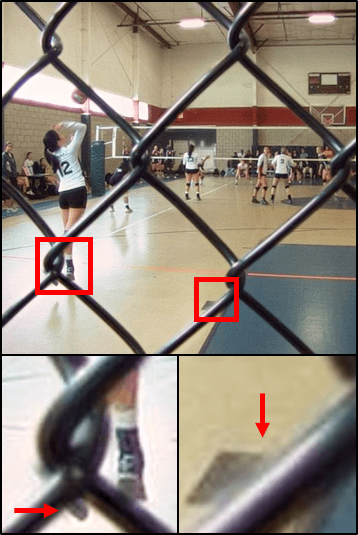}}%
  \subfigure[SR]{\label{fig:subfig:44}
    \includegraphics[width=0.19\linewidth]{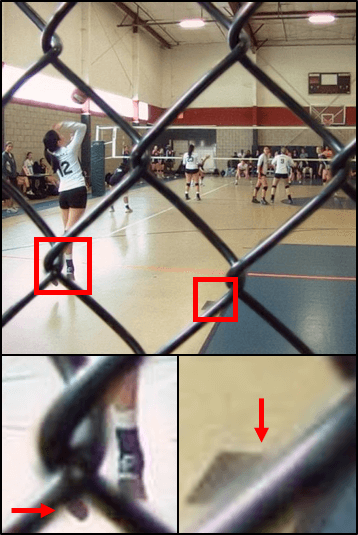}}%
  \subfigure[NSCT-SR]{\label{fig:subfig:45}
    \includegraphics[width=0.19\linewidth]{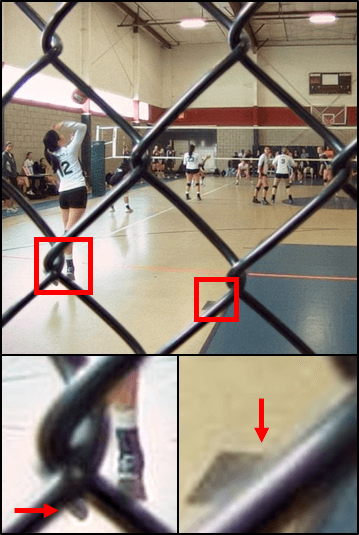}}\\
  \subfigure[GF]{\label{fig:subfig:46}
    \includegraphics[width=0.19\linewidth]{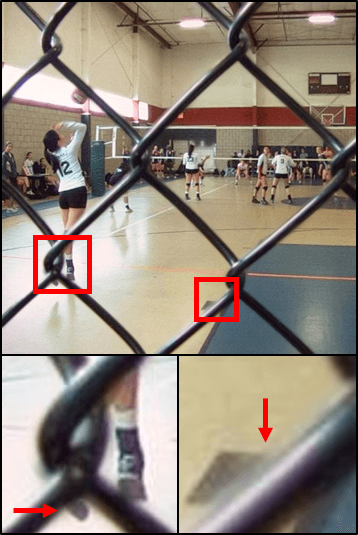}}
  \subfigure[DSIFT]{\label{fig:subfig:47}
    \includegraphics[width=0.19\linewidth]{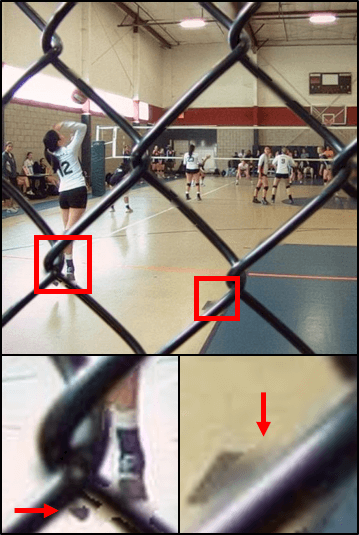}}%
  \subfigure[CNN]{\label{fig:subfig:48}
    \includegraphics[width=0.19\linewidth]{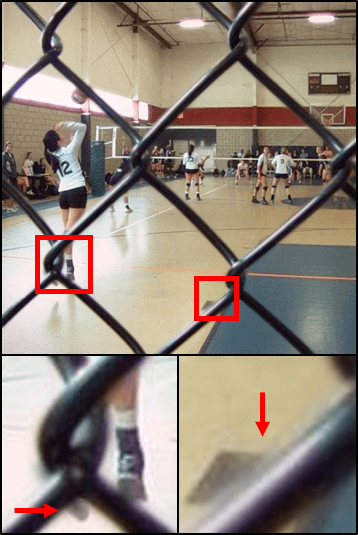}}%
  \subfigure[BA-Fusion]{\label{fig:subfig:49}
    \includegraphics[width=0.19\linewidth]{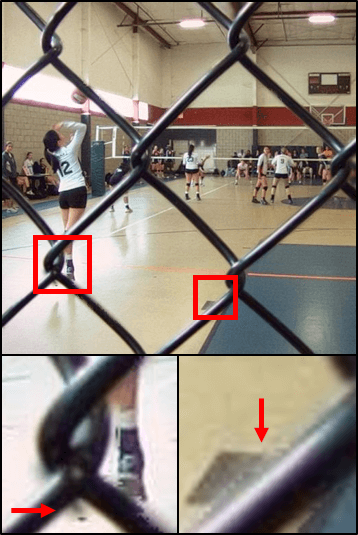}}%
  \subfigure[MMF-Net]{\label{fig:subfig:410}
    \includegraphics[width=0.19\linewidth]{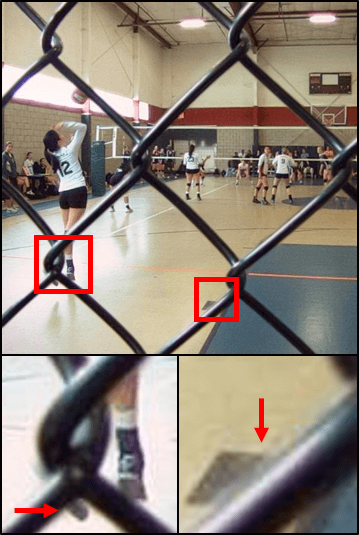}}
  \caption{The fusion results of different methods on `Lytro-05'. Compared with other MFIF methods, both the grille's edge in the foreground and the people as well as the floor in the background are clearer in our MMF-Net's result, as shown in the enlarged squares.}
\label{fig:4} 
\end{figure*}

\begin{figure*}[htbp]
\centering 
  \subfigure[Source A]{\label{fig:subfig:51}
    \includegraphics[width=0.19\linewidth]{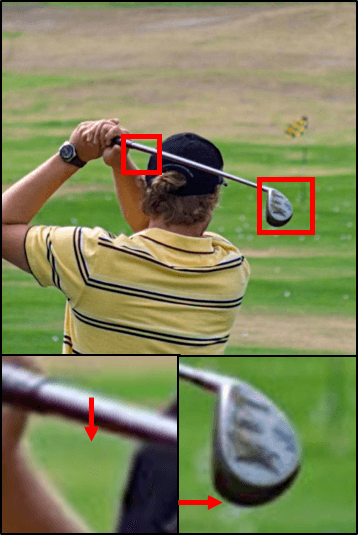}}%
  \subfigure[Source B]{\label{fig:subfig:52}
    \includegraphics[width=0.19\linewidth]{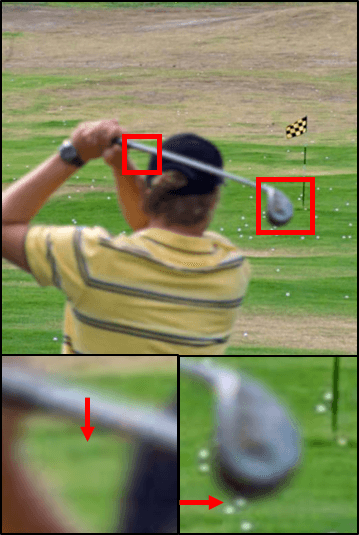}}
  \subfigure[NSCT]{\label{fig:subfig:53}
    \includegraphics[width=0.19\linewidth]{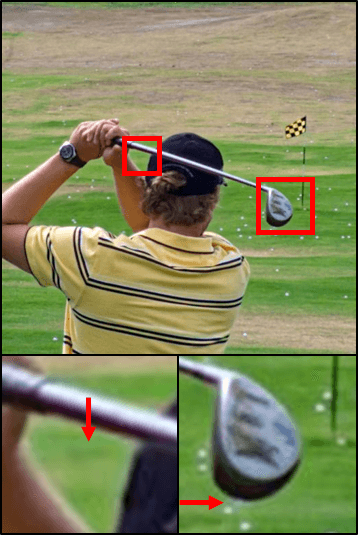}}%
  \subfigure[SR]{\label{fig:subfig:54}
    \includegraphics[width=0.19\linewidth]{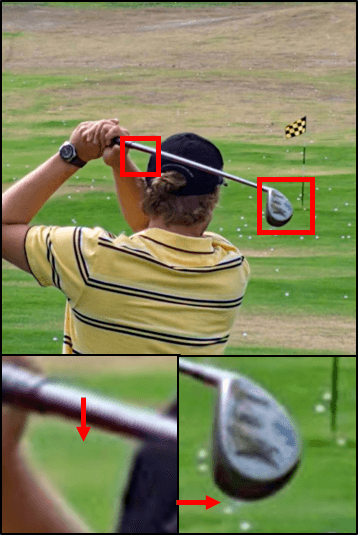}}%
  \subfigure[NSCT-SR]{\label{fig:subfig:55}
    \includegraphics[width=0.19\linewidth]{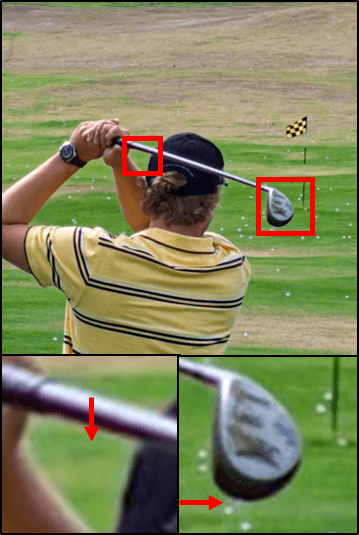}}\\
  \subfigure[GF]{\label{fig:subfig:56}
    \includegraphics[width=0.19\linewidth]{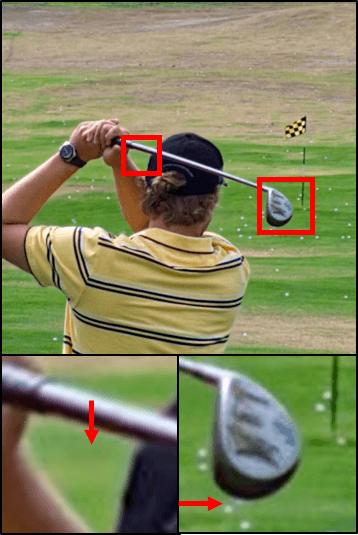}}
  \subfigure[DSIFT]{\label{fig:subfig:57}
    \includegraphics[width=0.19\linewidth]{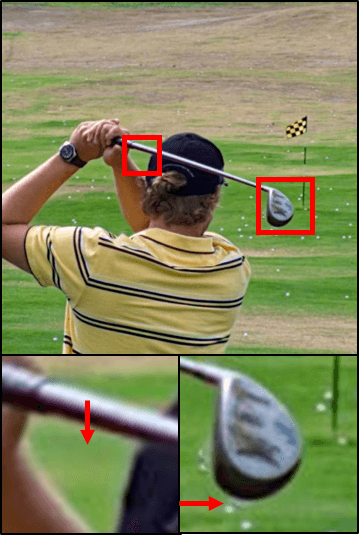}}%
  \subfigure[CNN]{\label{fig:subfig:58}
    \includegraphics[width=0.19\linewidth]{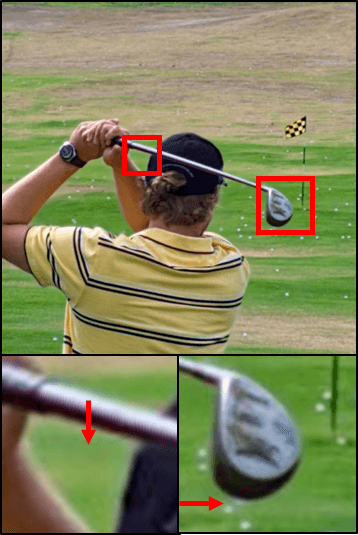}}%
  \subfigure[BA-Fusion]{\label{fig:subfig:59}
    \includegraphics[width=0.19\linewidth]{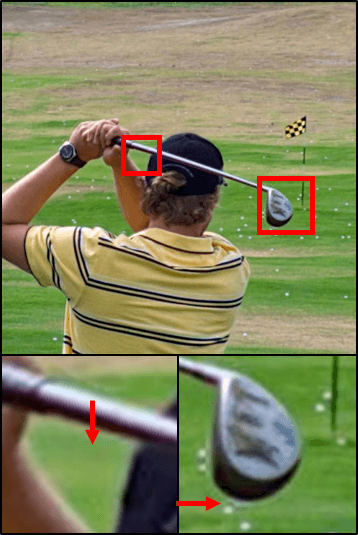}}%
  \subfigure[MMF-Net]{\label{fig:subfig:510}
    \includegraphics[width=0.19\linewidth]{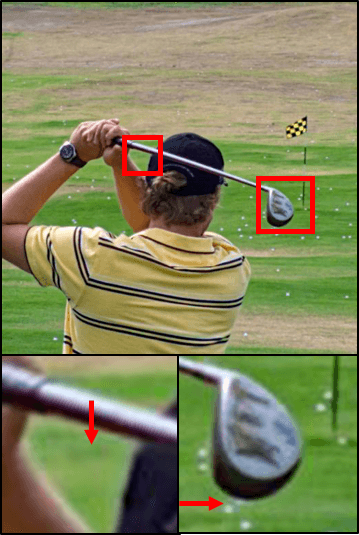}}
  \caption{The fusion results of different methods on `Lytro-01'. Compared with other MFIF methods, both the edge of golf club in the foreground and the golf balls as well as the grassland in the background are clearer in our MMF-Net's result, as shown in the enlarged squares.}
\label{fig:5} 
\end{figure*}

We collect 200 foreground images from datasets of matting \cite{AlphaMating, xu2017deep} with corresponding matte maps, and choose 1,200 background pictures from the COCO dataset \cite{COCO}. The background pictures are first resized to 512 $\times$ 512. Then for every foreground images, 20 background images are randomly chosen. The order of source images are random with probability equals 0.5, therefore, 4,000 image pairs are obtained in total.

\subsection{Training Settings}

For the training process, 4 $\times$ 1080Ti GPUs are used; and the test is carried on with a single GPU. The Adam solver is used with parameters $\beta_1$ = 0.9, $\beta_2$ = 0.999, and $\epsilon$ = $10^{-8}$. The batch size is set to 32, with the learning rate is set to 0.001. The model is trained on the generated dataset for 80 epochs. During the test process, it takes 0.27 seconds on average to fuse an image pairs of size 520 $\times$ 520.

\subsection{Comparison Settings}

We compare the proposed MMF-Net with 7 other multi-focus fusion methods: conventional methods including NSCT \cite{ZHANG20091334}, SR \cite{RN35}, NSCT-SR \cite{LIU2015147}, GF \cite{RN80} and DSIFT \cite{RN43}; network approaches including DCNN \cite{RN497}; and our previous work BA-Fusion \cite{MA2019125}. The experiments is conducted on dataset `Lytro', which is commonly used for MFIF.

Four widely used objective metrics to assess fusion image quality are used to evaluate the results \cite{zhao2018multi}: average gradient ($AG$) \cite{cui2015detail}, linear index of fuzziness ($LIF$) \cite{bai2012noise}, mean square deviation ($MSD$) and gray level difference ($GLD$). Their formulations are described as follows.

1) $LIF$: $LIF$ is an evaluation metric which can evaluate the enhancement of fused images:
\begin{align}
    LIF &= \frac{2}{MN}\sum_{m=1}^{M}\sum_{n=1}^{N}{\min\{p_{mn},(1-p_{mn})\}},\\
p_{mn}&=\sin[\frac{\pi}{2}(1-\frac{I(m,n)}{I_{max}})],
\end{align}
where $I(m,n)$ is the intensity of pixel $(m,n)$ in image $I$, and $I_{max}$ is the maximum intensity of image $I$.
A small $LIF$ indicates that the enhancement of fused image is good.

2) $AG$: $AG$ is a metric which uses gradient information to measure the quality of fused images:
\begin{equation}
 \begin{split}
     AG =& \frac{1}{(M-1)(N-1)} \times\\
    &\sum_{m=1}^{M-1}\sum_{n=1}^{N-1}\frac{1}{4} \sqrt{(\frac{\partial I(m,n)}{\partial m})^2+(\frac{\partial I(m,n)}{\partial n})^2},
 \end{split}
\end{equation}
where $\frac{\partial I(m,n)}{\partial m}$ and $\frac{\partial I(m,n)}{\partial n}$ are the gradients of the image in horizontal and vertical directions, respectively. A larger $AG$ means that the boundaries of the fused image are clearer.

3) $MSD$: $MSD$ measures image detail richness by calculating the difference between the intensity of each pixel and the mean intensity $\bar I$ of the fused image:
\begin{equation}
     MSD = \frac{1}{(M-1)(N-1)} 
    \sqrt{\sum_{m=1}^{M-1}\sum_{n=1}^{N-1}{(I(m,n)-\bar I)^2}}.
\end{equation}
A larger $MSD$ corresponds to a clearer fused image.

4) $GLD$: $GLD$ uses $L1$ norm to calculate the gradient information of the fused image:
\begin{equation}
 \begin{split}
  &GLD = \frac{1}{(M-1)(N-1)} \times\\
  &\sum_{m=1}^{M-1}\sum_{n=1}^{N-1} (|I(m,n)-I(m+1,n)|+|I(m,n)-I(m,n+1)|).
 \end{split}
\end{equation}
A lager $GLD$ indicates a fused image with clearer boundary.

\subsection{Experimental Results and Analysis}

\begin{table*}[htbp]
\centering
\caption{The quantitative comparison of different MFIF methods. For $AG$, $MSD$ and $GLD$, the larger values means better results; for $LIF$, the smaller values means better results. The best average result is in bold, and the number of image pairs (out of totally 20) that one method beats all the other methods is shown in the parentheses. Our proposed MMF-Net outperforms the other MFIF methods on all metrics.}\label{table11}
\begin{tabular}{l|rrrrrrrrr}
\hline
Metrics & NSCT & SR & NSCT-SR & GF & DSIFT & DCNN & BA-Fusion & MMF-Net \\
\hline
  $AG$  & 2.8750 (0)  & 2.8446 (0) & 2.8794 (0) & 2.8699 (0) & 2.9020 (4) & 2.8598 (0) & 2.9040 (5) & \textbf{2.9189} (11) \\
  $LIF$    & 0.4097 (1)  & 0.4083 (0) & 0.4093 (1) & 0.4081 (0) & 0.4075 (2) & 0.4080 (0) &  0.4075 (1) & \textbf{0.4071} (15) \\
  $MSD$  & 0.1108 (0)  & 0.1108 (0) & 0.1108 (1) & 0.1110 (1) & 0.1111 (1) & 0.1109 (0) &  0.1112 (0) & \textbf{0.1114} (17) \\
  $GLD$  & 14.2245 (0)  & 14.0740 (0) & 14.2467 (0) & 14.1954 (0) & 14.3400 (3) & 14.1460 (0) &  14.3550 (5) & \textbf{14.4388} (12) \\
\hline
\end{tabular}
\end{table*}

Fig.~\ref{fig:4} shows the visual comparison on an example `Lytro-05'  from the `Lytro' dataset. Compared with other approaches, the proposed MMF-Net performs well in general. Our results overcome the unclear boundary that exist in Figs.~\ref{fig:subfig:43}, \ref{fig:subfig:44}, \ref{fig:subfig:45}, \ref{fig:subfig:46} and \ref{fig:subfig:48}. In addition, the artifact area in Figs. \ref{fig:subfig:47} and \ref{fig:subfig:49} is not shown in our results. Moreover, both the grille's edge in foreground and the people as well as the floor in background are clearer in our result (Fig.~\ref{fig:subfig:410}), as shown in the enlarged squares.

Fig.~\ref{fig:5} shows the visual comparison on an example `Lytro-01'  from the `Lytro' dataset. Compared with other approaches, the proposed MMF-Net produces clearer fusion results, especially for the areas near the FDB. Our results are with the clear grassland which is not obtained in Figs.~\ref{fig:subfig:54}, \ref{fig:subfig:56}, \ref{fig:subfig:57}, \ref{fig:subfig:58} and \ref{fig:subfig:59}. And the golf ball under the boundary of the golf club are shown in our result, which is unclear in Figs. \ref{fig:subfig:53}, \ref{fig:subfig:54}, \ref{fig:subfig:55}, \ref{fig:subfig:57} and \ref{fig:subfig:59}. Moreover, both the edge of golf club in the foreground and the golf balls as well as the grassland in the background are clearer in our MMF-Net's result (Fig.~\ref{fig:subfig:510}), as shown in the enlarged squares.

The quantitative comparisons are shown in Table \ref{table11}. The larger values of metrics $AG$, $MSD$ and $GLD$ means better fusion results, while the smaller values of $LIF$ means better results. The results in Table \ref{table11} shown are the average values over the 20 pairs of test images, and the best average result of the compared methods is highlighted in bold. Among the 20 pairs, the number of image pairs that one method beats all the other methods is shown in the parentheses. As can be seen, the proposed MMF-Net remarkably outperforms the other fusion methods on all quality metrics.

\section{Conclusions}\label{s:conclusions}

In this paper, a cascaded boundary aware convolutional network called MMF-Net is proposed for multi-focus image fusion, along with a new $\alpha$-matte boundary defocus model. The proposed MMF-Net aims to solve the unclear areas near the focus/defocus boundary (FDB) in the fusion results, through two sub-nets to first generate a fusion guidance map and then refine the fusion results in the areas near the FDB. In addition, the dataset generated with the $\alpha$-matte model simulates the real-world images precisely, especially for the areas near the FDB. Experiments show that with the help of MMF-Net and the more realistic training data, the proposed MMF-Net outperforms the state-of-the-art methods both qualitatively and quantitatively.  

\appendices



\bibliographystyle{IEEEtran}
\bibliography{IEEEtran}

\begin{thebibliography}{10}
\providecommand{\url}[1]{#1}
\csname url@samestyle\endcsname
\providecommand{\newblock}{\relax}
\providecommand{\bibinfo}[2]{#2}
\providecommand{\BIBentrySTDinterwordspacing}{\spaceskip=0pt\relax}
\providecommand{\BIBentryALTinterwordstretchfactor}{4}
\providecommand{\BIBentryALTinterwordspacing}{\spaceskip=\fontdimen2\font plus
\BIBentryALTinterwordstretchfactor\fontdimen3\font minus
  \fontdimen4\font\relax}
\providecommand{\BIBforeignlanguage}[2]{{%
\expandafter\ifx\csname l@#1\endcsname\relax
\typeout{** WARNING: IEEEtran.bst: No hyphenation pattern has been}%
\typeout{** loaded for the language `#1'. Using the pattern for}%
\typeout{** the default language instead.}%
\else
\language=\csname l@#1\endcsname
\fi
#2}}
\providecommand{\BIBdecl}{\relax}
\BIBdecl

\bibitem{xu2017deep}
N.~Xu, B.~Price, S.~Cohen, and T.~Huang, ``Deep image matting,'' in
  \emph{Proceedings of the IEEE Conference on Computer Vision and Pattern
  Recognition}, 2017, pp. 2970--2979.

\bibitem{hassen2015objective}
R.~Hassen, Z.~Wang, and M.~M. Salama, ``Objective quality assessment for
  multiexposure multifocus image fusion,'' \emph{IEEE Transactions on Image
  Processing}, vol.~24, no.~9, pp. 2712--2724, 2015.

\bibitem{bouzos2019conditional}
O.~Bouzos, I.~Andreadis, and N.~Mitianoudis, ``Conditional random field model
  for robust multi-focus image fusion,'' \emph{IEEE Transactions on Image
  Processing}, vol.~28, no.~11, pp. 5636--5648, 2019.

\bibitem{kaur2015comparative}
P.~Kaur and M.~Kaur, ``A comparative study of various digital image fusion
  techniques: A review,'' \emph{International Journal of Computer
  Applications}, vol. 114, no.~4, pp. 26--31, 2015.

\bibitem{ZHANG20091334}
Q.~Zhang and B.~long Guo, ``Multifocus image fusion using the nonsubsampled
  contourlet transform,'' \emph{Signal Processing}, vol.~89, no.~7, pp. 1334 --
  1346, 2009.

\bibitem{RN35}
B.~Yang and S.~T. Li, ``Multifocus image fusion and restoration with sparse
  representation,'' \emph{IEEE Transactions on Instrumentation and
  Measurement}, vol.~59, no.~4, pp. 884--892, 2010.

\bibitem{zhang2016robust}
Q.~Zhang and M.~D. Levine, ``Robust multi-focus image fusion using multi-task
  sparse representation and spatial context,'' \emph{IEEE Transactions on Image
  Processing}, vol.~25, no.~5, pp. 2045--2058, 2016.

\bibitem{LIU2015147}
Y.~Liu, S.~P. Liu, and Z.~F. Wang, ``A general framework for image fusion based
  on multi-scale transform and sparse representation,'' \emph{Information
  Fusion}, vol.~24, pp. 147 -- 164, 2015.

\bibitem{RN85}
M.~Li, W.~Cai, and Z.~Tan, ``A region-based multi-sensor image fusion scheme
  using pulse-coupled neural network,'' \emph{Pattern Recognition Letters},
  vol.~27, no.~16, pp. 1948--1956, 2006.

\bibitem{RN43}
Y.~Liu, S.~P. Liu, and Z.~F. Wang, ``Multi-focus image fusion with dense
  {SIFT},'' \emph{Information Fusion}, vol.~23, pp. 139--155, 2015.

\bibitem{RN80}
S.~T. Li, X.~D. Kang, and J.~W. Hu, ``Image fusion with guided filtering,''
  \emph{IEEE Transactions on Image Processing}, vol.~22, no.~7, pp. 2864--2875,
  2013.

\bibitem{wang2010multi}
Z.~Wang, Y.~Ma, and J.~Gu, ``Multi-focus image fusion using pcnn,''
  \emph{Pattern Recognition}, vol.~43, no.~6, pp. 2003--2016, 2010.

\bibitem{RN91}
H.~Tang, B.~Xiao, W.~Li, and G.~Wang, ``Pixel convolutional neural network for
  multi-focus image fusion,'' \emph{Information Sciences}, vol. 433-434, pp.
  125--141, 2018.

\bibitem{RN497}
Y.~Liu, X.~Chen, H.~Peng, and Z.~Wang, ``Multi-focus image fusion with a deep
  convolutional neural network,'' \emph{Information Fusion}, vol.~36, pp.
  191--207, 2017.

\bibitem{8625482}
X.~{Guo}, R.~{Nie}, J.~{Cao}, D.~{Zhou}, L.~{Mei}, and K.~{He}, ``Fusegan:
  Learning to fuse multi-focus image via conditional generative adversarial
  network,'' \emph{IEEE Transactions on Multimedia}, vol.~21, no.~8, pp.
  1982--1996, 2019.

\bibitem{ma2019sesf}
B.~Ma, X.~Ban, H.~Huang, and Y.~Zhu, ``{SESF-Fuse}: An unsupervised deep model
  for multi-focus image fusion,'' \emph{arXiv preprint arXiv:1908.01703}, 2019.

\bibitem{yan2018unsupervised}
X.~Yan, S.~Z. Gilani, H.~Qin, and A.~Mian, ``Unsupervised deep multi-focus
  image fusion,'' \emph{arXiv preprint arXiv:1806.07272}, 2018.

\bibitem{zhao2018multi}
W.~Zhao, D.~Wang, and H.~Lu, ``Multi-focus image fusion with a natural
  enhancement via a joint multi-level deeply supervised convolutional neural
  network,'' \emph{IEEE Transactions on Circuits and Systems for Video
  Technology}, vol.~29, no.~4, pp. 1102--1115, 2018.

\bibitem{zhuo2011defocus}
S.~Zhuo and T.~Sim, ``Defocus map estimation from a single image,''
  \emph{Pattern Recognition}, vol.~44, no.~9, pp. 1852--1858, 2011.

\bibitem{Liu-TIP-2016}
S.~{Liu}, F.~{Zhou}, and Q.~{Liao}, ``Defocus map estimation from a single
  image based on two-parameter defocus model,'' \emph{IEEE Transactions on
  Image Processing}, vol.~25, no.~12, pp. 5943--5956, Dec 2016.

\bibitem{MA2019125}
H.~{Ma}, J.~{Zhang}, S.~{Liu}, and Q.~{Liao}, ``Boundary aware multi-focus
  image fusion using deep neural network,'' in \emph{2019 IEEE International
  Conference on Multimedia and Expo (ICME)}, July 2019, pp. 1150--1155.

\bibitem{he2010guided}
K.~He, J.~Sun, and X.~Tang, ``Guided image filtering,'' in \emph{European
  Conference on Computer Vision}.\hskip 1em plus 0.5em minus 0.4em\relax
  Springer, 2010, pp. 1--14.

\bibitem{chen2017robust}
Y.~Chen, J.~Guan, and W.-K. Cham, ``Robust multi-focus image fusion using edge
  model and multi-matting,'' \emph{IEEE Transactions on Image Processing},
  vol.~27, no.~3, pp. 1526--1541, 2017.

\bibitem{deng2009imagenet}
J.~Deng, W.~Dong, R.~Socher, L.-J. Li, K.~Li, and L.~Fei-Fei, ``{ImageNet}: A
  large-scale hierarchical image database,'' in \emph{Proceedings of the IEEE
  Conference on Computer Vision and Pattern Recognition}.\hskip 1em plus 0.5em
  minus 0.4em\relax IEEE, 2009, pp. 248--255.

\bibitem{RN618}
K.~M. He, X.~Y. Zhang, S.~Q. Ren, and J.~Sun, ``Deep residual learning for
  image recognition,'' in \emph{Proceedings of the IEEE Conference on Computer
  Vision and Pattern Recognition}, 2016, pp. 770--778.

\bibitem{AlphaMating}
C.~Rhemann, C.~Rother, J.~Wang, M.~Gelautz, P.~Kohli, and P.~Rott, ``A
  perceptually motivated online benchmark for image matting,'' in
  \emph{Proceedings of the IEEE Conference on Computer Vision and Pattern
  Recognition}, 2009, pp. 1826--1833.

\bibitem{COCO}
T.-Y. Lin, M.~Maire, S.~Belongie, J.~Hays, P.~Perona, D.~Ramanan,
  P.~Doll{\'a}r, and C.~L. Zitnick, ``{Microsoft COCO}: Common objects in
  context,'' in \emph{ECCV}, 2014, pp. 740--755.

\bibitem{cui2015detail}
G.~Cui, H.~Feng, Z.~Xu, Q.~Li, and Y.~Chen, ``Detail preserved fusion of
  visible and infrared images using regional saliency extraction and
  multi-scale image decomposition,'' \emph{Optics Communications}, vol. 341,
  pp. 199--209, 2015.

\bibitem{bai2012noise}
X.~Bai, F.~Zhou, and B.~Xue, ``Noise-suppressed image enhancement using
  multiscale top-hat selection transform through region extraction,''
  \emph{Applied optics}, vol.~51, no.~3, pp. 338--347, 2012.

\end{thebibliography}

%
%
%
%
%

\end{document}